\documentclass[conference]{IEEEtran}
\usepackage{cite}
\usepackage{amsmath,amssymb,amsfonts}
\usepackage{algorithmic}
\usepackage{graphicx}
\usepackage{textcomp}
\usepackage{caption}
\usepackage{makecell}
 \usepackage{multirow}
\usepackage{adjustbox}
\usepackage{comment}
\usepackage[ruled,vlined]{algorithm2e}
\SetKwInOut{Parameter}{Parameter}
\SetKwInOut{Input}{Input}
\SetKwInOut{Output}{Output}
\usepackage{pgfplots}
\usepackage{subcaption}
\ifCLASSINFOpdf
\else
\fi
\begin{document}
%
\title{Crafting Adversarial Examples for Deep Learning Based Prognostics (Extended Version)}

\author{\IEEEauthorblockN{Gautam Raj Mode, Khaza Anuarul Hoque}
\IEEEauthorblockA{Dept. of Electrical Engineering and Computer Science\\
University of Missouri, Columbia, MO, USA\\
gmwyc@mail.missouri.edu, hoquek@missouri.edu}
}


%


\maketitle

\begin{abstract}
In manufacturing, unexpected failures are considered a primary operational risk, as they can hinder productivity and can incur huge losses. State-of-the-art Prognostics and Health Management (PHM) systems incorporate Deep Learning (DL) algorithms and Internet of Things (IoT) devices to ascertain the health status of equipment, and thus reduce the downtime, maintenance cost and increase the productivity. Unfortunately, IoT sensors and DL algorithms, both are vulnerable to cyber attacks, and hence pose a significant threat to PHM systems. In this paper, we adopt the adversarial example crafting techniques from the computer vision domain and apply them to the PHM domain. Specifically, we craft adversarial examples using the Fast Gradient Sign Method (FGSM) and Basic Iterative Method (BIM) and apply them on the Long Short-Term Memory (LSTM), Gated Recurrent Unit (GRU), and Convolutional Neural Network (CNN) based PHM models. We evaluate the impact of adversarial attacks using NASA's turbofan engine dataset. The obtained results show that all the evaluated PHM models are vulnerable to adversarial attacks and can cause a serious defect in the remaining useful life estimation. The obtained results also show that the crafted adversarial examples are highly transferable and may cause significant damages to PHM systems.

\end{abstract}


%
\IEEEpeerreviewmaketitle

\section{Introduction}
The advent of Industry 4.0 in automation and data exchange leads us toward a constant evolution in smart manufacturing environments, including an intensive utilization of Internet-of-Things (IoT) and Deep Learning (DL). Specifically, the state-of-the-art Prognostics and Health Management (PHM) \cite{gan2020prognostics} has shown great success in achieving a competitive edge in Industry 4.0 by reducing the maintenance cost, downtime and increasing the productivity by making data-driven informed decisions. For instance, modern PHM techniques can help reduce downtime by 35\%-45\%, maintenance cost by 20\%-25\%, and can increase production by 20\%-25\%~\cite{ibm}. The ability to sense changes in the physical world (such as temperature, vibration, pressure, etc.) using IoT sensors, and to analyze the sensed data using the state-of-the-art DL algorithms for different prognostic tasks such as the Remaining Useful Life (RUL) prediction has enabled a highly reliable and cost-efficient industrial automation framework. Unfortunately, IoT sensors are also known for their vulnerability to cyber attacks~\cite{inrel8,inrel4}, and DL algorithms can also be easily fooled by adversarial examples~\cite{adrel2}. From the perspective of computer vision, an adversarial example can be an image formed by making small perturbations (insignificant to the human eye) so that a classifier misclassifies it with high confidence. The highly-connected IoT sensors and DL utilized in PHM systems tend to inherit their respective vulnerabilities, thus making them a lucrative target for cyber-attackers~\cite{risk1}. According to a recent report from the \emph{Malwarebytes}, cyber-threats against businesses/factories have increased by more than 200\% over the past year~\cite{malware}. Another interesting fact is that the adversarial examples can often transfer from one model to another model, which means that it is possible to attack models to which the attacker does not have access~\cite{adrel2}. Such adversarial attacks have been extensively studied in the computer vision domain~\cite{akhtar2018threat}. Even though advanced data-driven PHM depends on DL, it is very surprising that the impact of adversarial attacks on the PHM domain has not been studied yet. 

In manufacturing, adversarial attacks can lead to a wrong prognostic decision, e.g., a wrong estimation of RUL can delay the maintenance of a machine leading to unexpected failures. Such unexpected failures are considered a primary operational risk, as they can hinder productivity and can incur a huge loss. For example, in the modern automotive industry, an assembly line has several robots working on a car, and if even one robot fails, it will result in the halt of the entire assembly line, causing loss of valuable production time and increased production cost. In another situation, a wrong prognostic prediction in an operating autonomous vehicle, or aircraft may lead to loss of human lives. Thus, even though the utilization of IoT and ML is revolutionizing the smart industry, the vulnerabilities related to IoT and ML possesses a great challenge for Industry 4.0.

\textit{Contribution.} In our previous work~\cite{mode2020impact}, we showed that even a very small amount of randomly generated noise injected to the IoT sensors can greatly defect the RUL estimation. In this paper, we go beyond the concept of randomly generated noise for PHM. We adopt adversarial attacks from the computer vision domain, formalize them, craft adversarial examples for the PHM domain, and perform a quantitative analysis of their impacts. To be specific, we use the Fast Gradient Sign Method (FGSM) \cite{goodfellow2014explaining} and Basic Iterative Method (BIM) \cite{kurakin2016adversarial} to craft adversarial examples for Long Short-Term Memory (LSTM)~\cite{LSTM-org11}, Gated Recurrent Unit (GRU)~\cite{GRU-org}, and Convolutional Neural Network (CNN)~\cite{CNN-org} based PHM models. We perform an empirical study of adversarial attacks in real-life scenarios using NASA's C-MAPSS dataset \cite{Dataset}. We also perform a comprehensive study of the transferability property of adversarial examples in DL-based PHM models. \textit{To the knowledge of the authors, this is the first work that shows the impact of adversarial attacks on DL enabled PHM systems}.

\textit{Paper organization.} The rest of the paper is organized as follows. Section II gives an overview of IoT and DL enabled PHM architectures and their associate threats. Section III formalizes adversarial attacks on the DL-based prognostics and presents the algorithms for crafting FGSM and BIM adversarial examples. Section IV presents the experimental results showing the impact of adversarial examples on a PHM case study.  Finally, Section V concludes the paper.

\section{Background}
\label{sec:background}

In this section, we present an overview of IoT and DL enabled PHM architectures and their associate threats.

\begin{figure}[t!]
	\centering
	\includegraphics[width=0.5\textwidth]{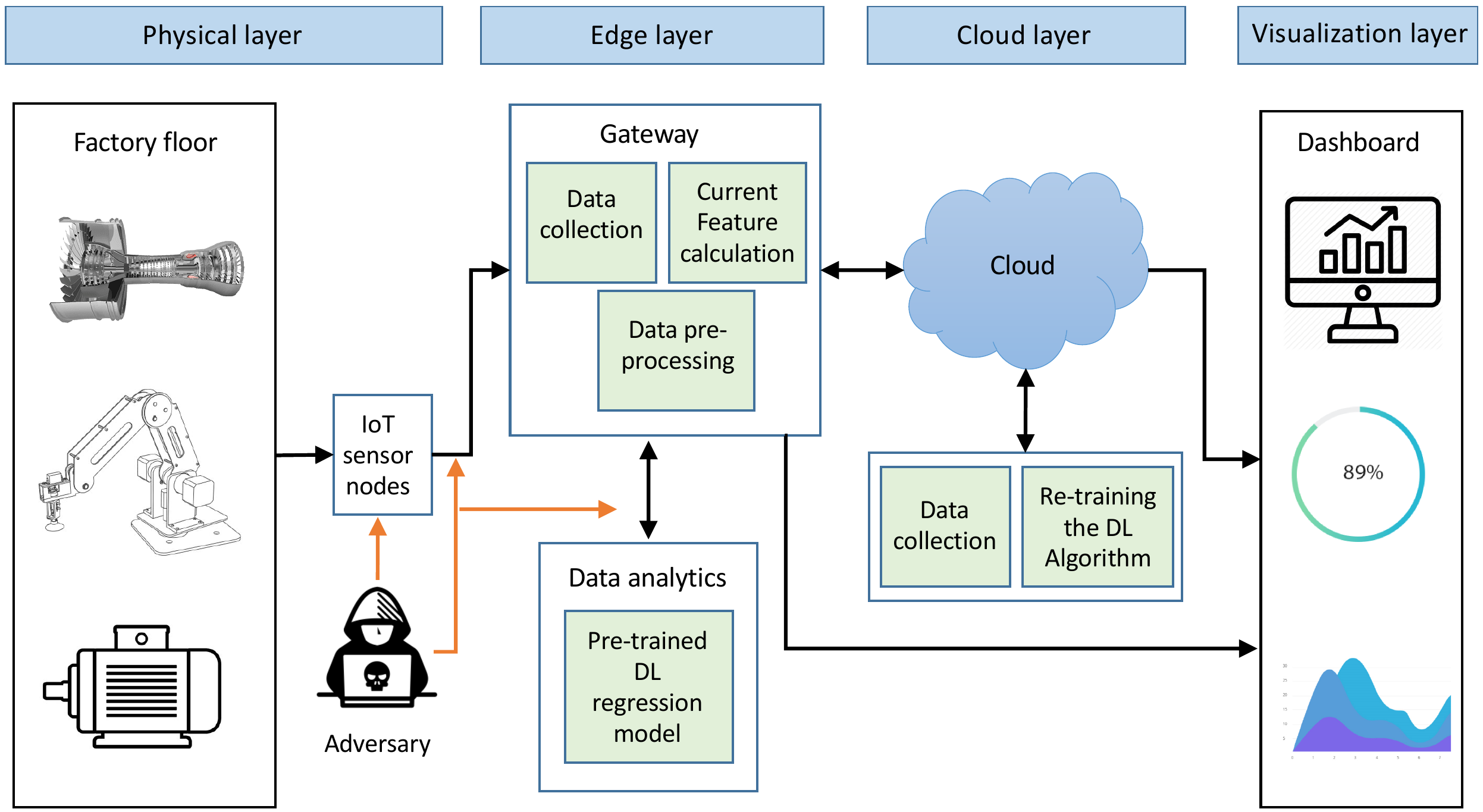}
	\caption{PHM cloud-edge architecture and threat model}
	\label{fig:cloudArch}
	
\end{figure}

\subsection{Internet of Things (IoT) and deep learning in prognostics}
Prognostics and Health Management (PHM) has gained a lot of attention in recent years for intelligent manufacturing in the context of Industry 4.0~\cite{zhang2019review}. The state-of-the-art PHM incorporates both IoT devices for sensing~\cite{gan2020prognostics}, and machine learning algorithms for analyzing the sensed data, and thus making a way for smart analytics to predict the future state of equipment. The equipment monitoring system includes input from IoT sensors that measure different parameters, such as pressure, temperature, speed, vibration, etc. Employing these IoT sensors eliminates the requirement for manual inspection/analysis, and hence the operating condition of equipment can be monitored using automated PHM systems. In PHM, Remaining Useful Lifetime (RUL) indicates the amount of time left before a piece of equipment or machine fails or degrades to a point at which it cannot perform its intended function anymore. Indeed, it is important to have an accurate RUL estimation since an early prediction may result in over-maintenance and a late prediction could lead to catastrophic failures. From a machine learning perspective, the prognostic is a regression problem, as the target value (RUL) is in the real domain. Thus, an RUL estimation involves learning a function that maps the condition of machines to its RUL estimates.

In the era of big data, an efficient way to analyze the huge amount of data is crucial. Deep learning (DL) provides a complementary approach for analyzing data obtained from IoT devices to provide accurate insights by identifying failure signatures, profiles, and providing an actionable prediction of failure through RUL estimation~\cite{zhang2019review,fink2020potential}. DL algorithms, especially LSTM, GRU, and CNN have shown great success in prognostics tasks~\cite{yang2019remaining,zhang2019remaining,park2020lstm,li2019directed,ren2019multi,chen2019gated}. \figurename \ref{fig:cloudArch} shows an overview of an IoT and DL enabled cloud-edge architecture, divided into physical, edge, cloud, and visualization layer~\cite{IoTLayersInfra}, and also a potential attack scenario. The \textit{physical layer} involves several IoT sensors nodes, which collect different sensor measurements from equipment. The data from the sensor nodes are sent to the edge for processing. The \textit{edge layer} pre-process the data for passing to the next \textit{cloud} layer for training/re-training purposes. The \textit{edge layer} also has a pre-trained deep learning model to predict the future health state of the equipment. \textit{The cloud} layer performs in-depth analysis and also performs training/re-training of the DL models on the newly arrived data from the edge layer. After the re-training, the re-trained DL model is sent back to the edge layer for improved accuracy of the machine's health state prediction. The \textit{visualization layer} uses the data collected from the field, along with the results from the PHM model, and provides a visual representation of actionable insights to an engineer.

\subsection{Vulnerabilities PHM systems}
\noindent \textbf{Attacks on IoT sensors:} IoT devices and deep learning have paved the way for smart analytics, however, their vulnerabilities also create new means of attacks. As mentioned earlier, in the PHM context, IoT sensors collect a variety of parameters including temperature, pressure, speed, vibration, etc., which is then transmitted via a network (in many cases using a wireless network) to a centralized processing unit to make informed decisions. The data collected from the sensors greatly influence the prognostics and maintenance related decisions. Even if the network is secured, any attacks on the sensors or the infrastructure could result in incorrect decisions and would result in a considerable impact on the performance of the PHM system~\cite{sadeghi2015security, IoT2}. For instance, attackers can use the sensors to transfer malicious code, trigger messages to activate a malware planted in an IoT device~\cite{inrel5}, capture sensitive information shared between devices~\cite{inrel8,inrel9}, or even capture encryption and depreciation keys for extracting encrypted information~\cite{inrel10}. Understanding these sensor-based threats is necessary for researchers to design reliable solutions to detect and prevent these threats efficiently. As a result, IoT sensor-based threats have gained a lot of attention from researchers in academia and in industry \cite{IoT3, sadeghi2015security, IoT2}. Unfortunately, most of the existing IoT security frameworks are not suitable for detecting sensor-based threats at the system level~\cite{inrel4}. IoT sensors are typically small and hence they limited by power and resource constraints. This is indeed an obstacle for implementing the complicated security mechanisms on IoT sensors and devices. Note, sensor attacks detection and mitigation is also an active research problem in the cyber-physical system domain~\cite{ding2018survey, shoukry2017secure}. However, in the context of industrial automation, how attacks on IoT sensors influence ML algorithms for making incorrect decisions is yet to be explored in detail.

\vspace{0.5em}
\noindent \textbf{Adversarial attacks on deep learning:} In addition to IoT attacks, the performance of a PHM system can also be considerably affected by orchestrated security attacks on DL algorithms. The study of the effect of adversarial attacks on machine learning techniques is known as \emph{adversarial machine learning}, and is one of the most active research topics in the deep learning community~\cite{adversarial2}. In~\cite{adrel2, kurakin2016adversarial, adrel5}, the authors have shown how adversarial examples can be used for deep learning algorithms to make a wrong classification. Since attacks on DL algorithms may have catastrophic consequences, their detection and mitigation have been explored in many recent literature~\cite{mitrel1, mitrel2}. However, most of them are venerable to future attacks~\cite{carlini2017provably}. For instance, in \cite{ganin2016domain, kurakin2018ensemble}, the authors explored the adversarial training technique to mitigate the adversarial attacks. Adversarial training injects adversarial examples into training data to increase robustness. Unfortunately, later on, researchers found the adversarial training technique to be inefficient for mitigating those attacks~\cite{zhang2019limitations}. A detailed survey of proposed defenses in the adversarial ML domain can be found in~\cite{chakraborty2018adversarial}. Even though adversarial machine learning is an active research area, most of the works in this area are limited to the computer vision domain or its variants. The effect of adversarial attacks in other domains, such as regression tasks for PHM is not yet explored. In our work, we study the impact of adversarial attacks on DL based regression models for PHM. 

\vspace{0.5em}
\noindent \textbf{PHM attack scenarios:} Let us consider an attack scenario as shown in the \figurename~\ref{fig:cloudArch}. An adversary can either compromise the physical sensors or the communication network used between IoT sensor nodes and the edge. If successful, an adversary can capture the sensor data, crafting adversarial examples for the captured data, and inject the crafted adversarial examples into the PHM system through a False Data Injection (FDI)~\cite{rahman2012false} attack. Indeed, the cloud layer also has its vulnerabilities~\cite{kumar2019cloud}, however, typically third-party cloud services such as AWS, Azure, etc. have their security measures~\cite{backes2019one,de2019azure} which makes them less vulnerable when compared to the physical and edge layer.

\section{Adversarial Attacks on Prognostics}
\vspace{-1mm}
In this section, we formalize the adversarial attacks in the PHM domain and present the adversarial example generation algorithms. 
\subsection{Formalization of the problem}
A machine or a piece of equipment in a PHM system has several sensors that record different parameters. These sensor measurements are recorded at every time step and hence constitute for multivariate time-series data~\cite{mode2020adversarial}. 

\noindent \emph{Definition 1:} Let $M$ be a multivariate time-series (MTS). Assuming there are $N$ sensors in an equipment, the multi-variate time-series can be defined as a sequence such that $M = [m_1,m_2,...,m_T]$, $T=\mid M \mid$ is the length of $M$, and $m_i\in\mathbb{R}^N$ is a $N$ dimension data point at time $i\in[1,T]$ representing the sensor measurements.

\noindent \emph{Definition 2:} $D = {(m_1,RUL_1), (m_2,RUL_2),...,(m_T,RUL_T)}$ is the dataset of pairs $(m_i,RUL_i)$ where $RUL_i$ is a label (RUL value at that time instant) corresponding to $m_i$.

\noindent \emph{Definition 3:} Time series regression task consists of training the model on $D$ in order to predict $\hat{RUL}$ from the possible inputs. Let $f(\cdot):\mathbb{R}^{T \times N} \rightarrow \hat{RUL}$ represent a DL for regression. $J_f(\cdot,\cdot)$ denotes the cost function of the model $f$.

\noindent \emph{Definition 4:} ${M}'$ denotes the adversarial example, a perturbed version of $M$ such that $\hat{RUL} \neq \hat{RUL}'$ and $\left \| M - {M}' \right \| \leq \epsilon $. where $\epsilon \geq 0 \in \mathbb{R}$ is a maximum perturbation magnitude.

Given a trained DL model $f$ and an input MTS $M$, crafting an adversarial example $M'$ can be described as a box-constrained optimization problem.
\begin{gather*}
\min_{M'} {\left \| {M}'- M \right \|} ~s.t.\\
f(M^{'})=\hat{RUL}',~ f(M)=\hat{RUL}~ and~ \hat{RUL} \neq \hat{RUL}'  
\end{gather*}





\begin{algorithm}[t]
\SetAlgoLined
\Input{Original multivariate time series $M$ from an equipment and its corresponding label $\hat{RUL}$}
\Output{Perturbed multivariate time series $M'$}
\Parameter{$\epsilon$}
 $\eta =\epsilon \cdot sign(\triangledown_mJ_f(M,\hat{RUL}))$\;
  $M'= M + \eta$\;
 \caption{FGSM algorithm for PHM} \label{alg:fgsm}
\end{algorithm}

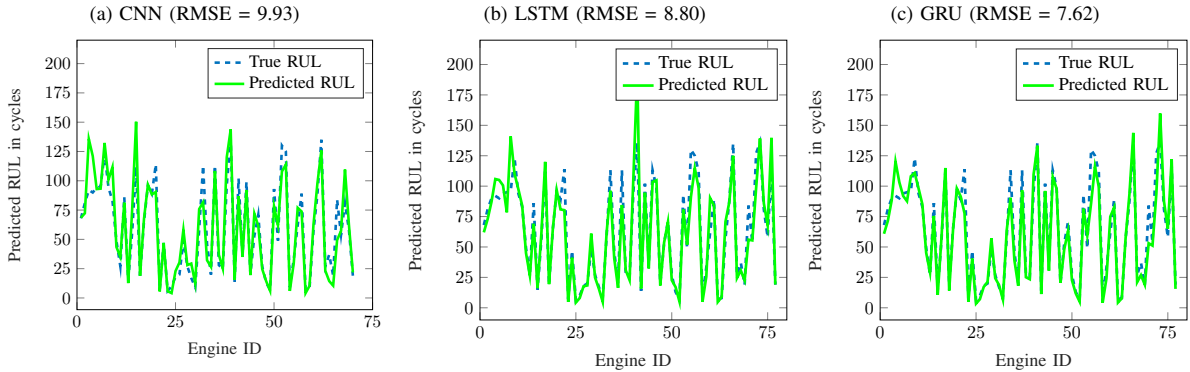
\begin{figure*}[t]
\centering
\resizebox{0.88\textwidth}{!}{
	\begin{subfigure}[t]{.33\textwidth}
		\centering
		\caption{CNN (RMSE = 9.93)}
		{\resizebox{\textwidth}{!}{
%
%
\definecolor{mycolor2}{rgb}{0.00000,0.44700,0.74100}%
\definecolor{mycolor1}{rgb}{0.85000,0.32500,0.09800}%
\definecolor{mycolor3}{rgb}{0,128,0}%
\definecolor{mycolor4}{rgb}{1.0, 0.75, 0.0}%
\begin{tikzpicture}

\begin{axis}[%
width=\textwidth,
height=2.15in,
at={(2.239in,0.602in)},
legend pos=south east,
scale only axis,
xmin=0,
xmax=75,
xlabel style={font=\color{white!15!black}},
xlabel={Engine ID},
ymin=-10,
ymax=220,
xtick ={0,25,50,75},
ylabel style={font=\color{white!15!black}},
ylabel={Predicted RUL in cycles},
ytick ={0,25,50,75,100,125,150,175,200},
axis background/.style={fill=white},
legend style={legend cell align=left, align=left,draw=white!15!black},
legend pos=north east
]
\addplot [color=mycolor2,dashed,line width = 1.5pt]
  table[row sep=crcr]{%
1	68	\\
2	81	\\
3	92	\\
4	90	\\
5	94	\\
6	95	\\
7	123	\\
8	94	\\
9	83	\\
10	49	\\
11	27	\\
12	86	\\
13	15	\\
14	56	\\
15	112	\\
16	19	\\
17	65	\\
18	96	\\
19	89	\\
20	114	\\
21	7	\\
22	47	\\
23	6	\\
24	10	\\
25	18	\\
26	20	\\
27	49	\\
28	27	\\
29	17	\\
30	9	\\
31	58	\\
32	113	\\
33	46	\\
34	20	\\
35	113	\\
36	28	\\
37	25	\\
38	96	\\
39	136	\\
40	14	\\
41	102	\\
42	36	\\
43	99	\\
44	20	\\
45	53	\\
46	71	\\
47	27	\\
48	13	\\
49	7	\\
50	93	\\
51	49	\\
52	130	\\
53	125	\\
54	9	\\
55	33	\\
56	62	\\
57	89	\\
58	7	\\
59	8	\\
60	57	\\
61	88	\\
62	135	\\
63	27	\\
64	37	\\
65	19	\\
66	84	\\
67	54	\\
68	81	\\
69	58	\\
70	19	\\
};
\addlegendentry{True RUL}

\addplot [color=mycolor3,line width=1.5pt]
  table[row sep=crcr]{%
1	68.43399	\\
2	72.28472	\\
3	135.70264	\\
4	121.11835	\\
5	92.90427	\\
6	93.268105	\\
7	132.32799	\\
8	100.97358	\\
9	111.68175	\\
10	43.534203	\\
11	35.083103	\\
12	81.33275	\\
13	12.6685705	\\
14	75.47095	\\
15	150.4011	\\
16	19.066868	\\
17	67.504364	\\
18	96.62007	\\
19	88.308685	\\
20	90.458244	\\
21	5.572108	\\
22	46.984913	\\
23	5.688321	\\
24	4.1292844	\\
25	22.93711	\\
26	30.311783	\\
27	58.247066	\\
28	28.05115	\\
29	29.3678	\\
30	12.553413	\\
31	75.849846	\\
32	81.64887	\\
33	32.159946	\\
34	26.604113	\\
35	108.36444	\\
36	36.83136	\\
37	25.084362	\\
38	120.04944	\\
39	143.86647	\\
40	15.364826	\\
41	86.90913	\\
42	34.969227	\\
43	92.10425	\\
44	19.494314	\\
45	72.00111	\\
46	61.445213	\\
47	24.65117	\\
48	14.332393	\\
49	6.129045	\\
50	78.13131	\\
51	62.056744	\\
52	107.96617	\\
53	113.87377	\\
54	6.004906	\\
55	29.592495	\\
56	76.946144	\\
57	73.76839	\\
58	4.142157	\\
59	10.319335	\\
60	62.35982	\\
61	97.7136	\\
62	125.848816	\\
63	23.30614	\\
64	14.374791	\\
65	10.534246	\\
66	49.36004	\\
67	57.2081	\\
68	109.59235	\\
69	62.462265	\\
70	21.788422	\\
};
\addlegendentry{Predicted RUL}

\end{axis}
\end{tikzpicture}%
}\label{fig:tempSe}}
	\end{subfigure}~
	\begin{subfigure}[t]{.33\textwidth}
		\centering
		\caption{LSTM (RMSE = 8.80)}
		{\resizebox{\textwidth}{!}{
%
%
\definecolor{mycolor2}{rgb}{0.00000,0.44700,0.74100}%
\definecolor{mycolor1}{rgb}{0.85000,0.32500,0.09800}%
\definecolor{mycolor3}{rgb}{0,128,0}%
\definecolor{mycolor4}{rgb}{1.0, 0.75, 0.0}%
\begin{tikzpicture}

\begin{axis}[%
width=\textwidth,
height=2.15in,
at={(2.239in,0.602in)},
legend pos=south east,
scale only axis,
xmin=0,
xmax=80,
xlabel style={font=\color{white!15!black}},
xlabel={Engine ID},
ymin=-10,
ymax=220,
xtick ={0,25,50,75},
ylabel style={font=\color{white!15!black}},
ylabel={Predicted RUL in cycles},
ytick ={0,25,50,75,100,125,150,175,200,225,250},
axis background/.style={fill=white},
legend style={legend cell align=left, align=left,draw=white!15!black},
legend pos=north east
]
\addplot [color=mycolor2,dashed,line width = 1.5pt]
  table[row sep=crcr]{%
1	68	\\
2	81	\\
3	90	\\
4	92	\\
5	90	\\
6	94	\\
7	95	\\
8	96	\\
9	123	\\
10	94	\\
11	83	\\
12	49	\\
13	27	\\
14	86	\\
15	15	\\
16	56	\\
17	112	\\
18	19	\\
19	65	\\
20	96	\\
21	89	\\
22	114	\\
23	7	\\
24	47	\\
25	6	\\
26	10	\\
27	18	\\
28	20	\\
29	49	\\
30	27	\\
31	17	\\
32	9	\\
33	58	\\
34	113	\\
35	46	\\
36	20	\\
37	113	\\
38	28	\\
39	25	\\
40	96	\\
41	136	\\
42	14	\\
43	102	\\
44	36	\\
45	113	\\
46	99	\\
47	20	\\
48	53	\\
49	71	\\
50	27	\\
51	13	\\
52	7	\\
53	93	\\
54	49	\\
55	130	\\
56	125	\\
57	112	\\
58	9	\\
59	33	\\
60	62	\\
61	89	\\
62	7	\\
63	8	\\
64	57	\\
65	88	\\
66	135	\\
67	27	\\
68	37	\\
69	19	\\
70	84	\\
71	54	\\
72	127	\\
73	136	\\
74	81	\\
75	58	\\
76	116	\\
77	19	\\
};
\addlegendentry{True RUL}

\addplot [color=mycolor3,line width=1.5pt]
  table[row sep=crcr]{%
1	62.01703	\\
2	72.96454	\\
3	87.12062	\\
4	105.84158	\\
5	105.065544	\\
6	100.61328	\\
7	78.461136	\\
8	140.92393	\\
9	110.335526	\\
10	97.57628	\\
11	83.99304	\\
12	45.01	\\
13	26.544094	\\
14	69.055336	\\
15	16.143446	\\
16	60.698143	\\
17	119.98399	\\
18	19.41734	\\
19	65.149826	\\
20	96.67172	\\
21	80.87248	\\
22	80.247406	\\
23	4.9050274	\\
24	41.198307	\\
25	4.363244	\\
26	8.012569	\\
27	17.663456	\\
28	18.38966	\\
29	60.999737	\\
30	24.845757	\\
31	15.6207	\\
32	4.8645906	\\
33	61.304733	\\
34	96.254395	\\
35	40.702435	\\
36	16.406265	\\
37	84.502655	\\
38	30.8285	\\
39	22.7753	\\
40	107.57002	\\
41	194.05489	\\
42	15.838239	\\
43	94.62038	\\
44	32.343967	\\
45	103.43178	\\
46	104.79587	\\
47	18.336033	\\
48	53.57677	\\
49	70.483055	\\
50	22.70794	\\
51	17.001968	\\
52	5.1743164	\\
53	80.737015	\\
54	51.916645	\\
55	91.316376	\\
56	114.7567	\\
57	98.73093	\\
58	4.867989	\\
59	25.499502	\\
60	90.24559	\\
61	84.09683	\\
62	4.8152127	\\
63	8.783452	\\
64	70.73162	\\
65	86.80951	\\
66	124.99324	\\
67	24.457417	\\
68	30.692822	\\
69	22.531078	\\
70	55.996403	\\
71	55.407444	\\
72	98.83301	\\
73	139.17036	\\
74	85.97555	\\
75	67.78777	\\
76	139.73529	\\
77	19.10324	\\
};
\addlegendentry{Predicted RUL}

\end{axis}
\end{tikzpicture}%
}\label{fig:tempS3}}
	\end{subfigure}~
	\begin{subfigure}[t]{.33\textwidth}
		\centering
		\caption{GRU (RMSE = 7.62)}
		{\resizebox{\textwidth}{!}{
%
%
\definecolor{mycolor2}{rgb}{0.00000,0.44700,0.74100}%
\definecolor{mycolor1}{rgb}{0.85000,0.32500,0.09800}%
\definecolor{mycolor3}{rgb}{0,128,0}%
\definecolor{mycolor4}{rgb}{1.0, 0.75, 0.0}%
\begin{tikzpicture}

\begin{axis}[%
width=\textwidth,
height=2.15in,
at={(2.239in,0.602in)},
legend pos=south east,
scale only axis,
xmin=0,
xmax=80,
xlabel style={font=\color{white!15!black}},
xlabel={Engine ID},
ymin=-10,
ymax=220,
xtick ={0,25,50,75},
ylabel style={font=\color{white!15!black}},
ylabel={Predicted RUL in cycles},
ytick ={0,25,50,75,100,125,150,175,200,225,250},
axis background/.style={fill=white},
legend style={legend cell align=left, align=left,draw=white!15!black},
legend pos=north east
]
\addplot [color=mycolor2,dashed,line width = 1.5pt]
  table[row sep=crcr]{%
1	68	\\
2	81	\\
3	90	\\
4	92	\\
5	90	\\
6	94	\\
7	95	\\
8	96	\\
9	123	\\
10	94	\\
11	83	\\
12	49	\\
13	27	\\
14	86	\\
15	15	\\
16	56	\\
17	112	\\
18	19	\\
19	65	\\
20	96	\\
21	89	\\
22	114	\\
23	7	\\
24	47	\\
25	6	\\
26	10	\\
27	18	\\
28	20	\\
29	49	\\
30	27	\\
31	17	\\
32	9	\\
33	58	\\
34	113	\\
35	46	\\
36	20	\\
37	113	\\
38	28	\\
39	25	\\
40	96	\\
41	136	\\
42	14	\\
43	102	\\
44	36	\\
45	113	\\
46	99	\\
47	20	\\
48	53	\\
49	71	\\
50	27	\\
51	13	\\
52	7	\\
53	93	\\
54	49	\\
55	130	\\
56	125	\\
57	112	\\
58	9	\\
59	33	\\
60	62	\\
61	89	\\
62	7	\\
63	8	\\
64	57	\\
65	88	\\
66	135	\\
67	27	\\
68	37	\\
69	19	\\
70	84	\\
71	54	\\
72	127	\\
73	136	\\
74	81	\\
75	58	\\
76	116	\\
77	19	\\
};
\addlegendentry{True RUL}

\addplot [color=mycolor3,line width=1.5pt]
  table[row sep=crcr]{%
1	60.850758	\\
2	71.45392	\\
3	88.32849	\\
4	120.17257	\\
5	103.18149	\\
6	91.96608	\\
7	87.659134	\\
8	108.14589	\\
9	110.857735	\\
10	98.26723	\\
11	85.10524	\\
12	45.193714	\\
13	29.073078	\\
14	75.5222	\\
15	10.573164	\\
16	58.192577	\\
17	114.95574	\\
18	14.167494	\\
19	64.13459	\\
20	98.06535	\\
21	91.49428	\\
22	77.61254	\\
23	5.0302105	\\
24	40.452652	\\
25	3.4699638	\\
26	7.04605	\\
27	17.148119	\\
28	21.326124	\\
29	57.121006	\\
30	24.3741	\\
31	18.648232	\\
32	6.835832	\\
33	63.83042	\\
34	90.46167	\\
35	39.096188	\\
36	18.086512	\\
37	96.64563	\\
38	25.142395	\\
39	23.508507	\\
40	109.61802	\\
41	133.2904	\\
42	11.279123	\\
43	96.09594	\\
44	30.471752	\\
45	107.61133	\\
46	97.916	\\
47	20.471214	\\
48	48.31953	\\
49	63.22254	\\
50	22.717337	\\
51	11.7796955	\\
52	5.1027827	\\
53	81.48596	\\
54	59.582993	\\
55	86.15106	\\
56	116.01097	\\
57	99.38782	\\
58	4.135732	\\
59	24.22692	\\
60	74.93103	\\
61	82.43964	\\
62	4.342657	\\
63	7.7642283	\\
64	57.76628	\\
65	93.731445	\\
66	143.81055	\\
67	21.049244	\\
68	26.870972	\\
69	19.014578	\\
70	53.08556	\\
71	51.145332	\\
72	93.54032	\\
73	159.78343	\\
74	87.31697	\\
75	64.99841	\\
76	122.211395	\\
77	15.534973	\\
};
\addlegendentry{Predicted RUL}
\end{axis}
\end{tikzpicture}%
}\label{fig:tempS4}}
	\end{subfigure}
	}
	\caption{Performance comparison of deep learning algorithms}\label{fig:LSTMStruct}
\end{figure*}

\subsection{Adversarial example generation for PHM}
We craft adversarial examples ($M'$) that defect the RUL predictions by increasing the cost of the model. In this work, we adopt and apply two adversarial example generation algorithms, Fast Gradient Sign Method (FGSM) \cite{goodfellow2014explaining} and Basic Iterative Method (BIM) \cite{kurakin2016adversarial}.\\

\begin{algorithm}[t]
\SetAlgoLined
\Input{Original multivariate time series $M$ from an equipment and its corresponding label $\hat{RUL}$}
\Output{Perturbed multivariate time series $M'$}
\Parameter{$I, \epsilon, \alpha$}
 $M' \leftarrow M$\;
 \While{$i=1 \leq I$}{
  $\eta =\alpha \cdot sign(\triangledown_mJ_f(M',\hat{RUL}))$\;
  $M'=M' + \eta$\;
  $M'=min \{M+\epsilon, max \{M-\epsilon,M'\}\}$\;
  $i++$\;
 }
 \caption{BIM algorithm for PHM} \label{alg:BIM}
 
\end{algorithm}

\noindent \textit{Fast Gradient Sign Method (FGSM)}: The FGSM was first proposed in \cite{goodfellow2014explaining} to generate adversarial images to fool the GoogLeNet model. The FGSM works by using the gradients of the neural network to create an adversarial example. This attack is also known as the one-shot method as the adversarial perturbation is generated by a single step computation. The attack is based on a one-step gradient update along the direction of the gradient's sign at each time step. This can be summarised using the following expression:
\begin{equation}
\eta =\epsilon \cdot sign(\triangledown_mJ_f(M,\hat{RUL}))
\end{equation} 

Here, $J_f$ is the cost function of model $f$, $\triangledown_m$ indicates the gradient of the model with respect to the original MTS $M$ with the correct label $\hat{RUL}$, $\epsilon$ denotes the hyper-parameter which controls the amplitude of the perturbation and $M'$ is adversarial MTS. Algorithm \ref{alg:fgsm} shows different steps of the FGSM attack.\\

\noindent \textit{Basic Iterative Method (BIM)}: The BIM \cite{kurakin2016adversarial} is an extension of FGSM. In BIM, FGSM is applied multiple times with small step size, and clipping is performed after each step to ensure that they are in the range [$M-\epsilon,M+\epsilon$] i.e. $\epsilon-neighbourhood$ of the original time series $M$. BIM is also known as Iterative-FGSM since FGSM is iterated with smaller step sizes. The adversarial examples generated through BIM are closer to the original input as perturbations are added iteratively and hence have a greater chance of fooling the network. Algorithm \ref{alg:BIM} shows different steps of the BIM attack. The algorithm requires three hyperparameters: 1. the per step small perturbation ($\alpha$); 2. the amount of maximum perturbation ($\epsilon$) and  3. the number of iterations ($I$). Here, the value of $\alpha$ is calculated using $\alpha=\epsilon/I$. Note, BIM does not rely on the approximation of the model, and the adversarial examples crafted through BIM are closer to the original samples when compared to FGSM. This is because the perturbations are added iteratively and hence have a better chance of fooling the network. However, compared to FGSM, BIM is computationally more expensive and slower. 


\begin{figure}[h]
	\centering
	\includegraphics[width=0.45\textwidth]{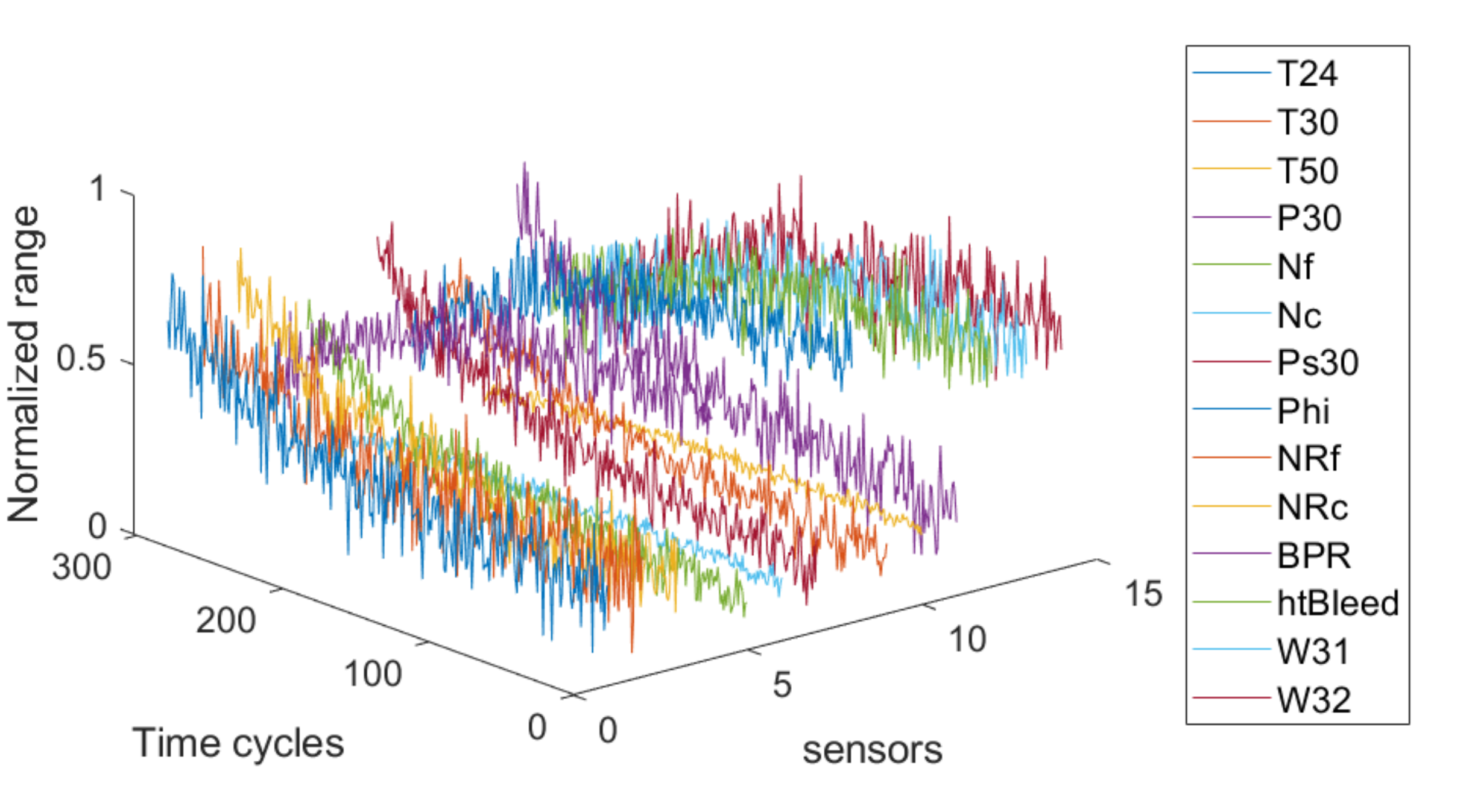}
	\caption{Normalized sensor measurements}
	\label{fig:3d}
	
\end{figure}

\begin{figure*}[t]
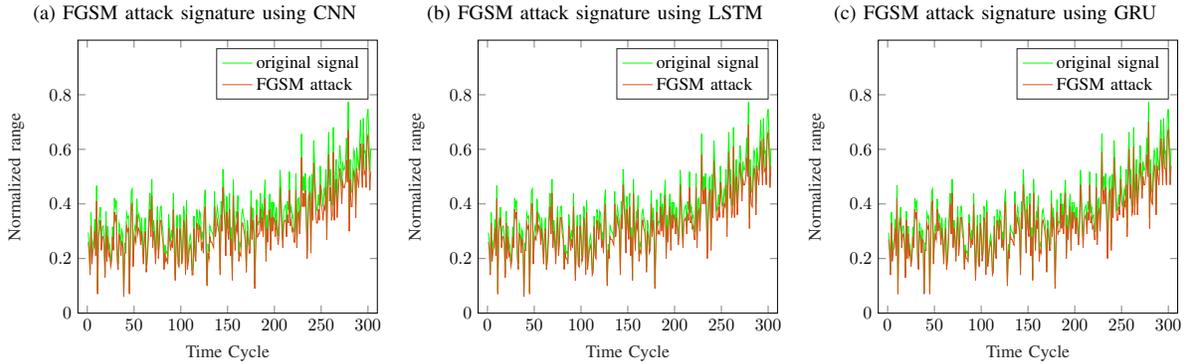

\centering
\resizebox{0.88\textwidth}{!}{
	\begin{subfigure}[t]{.33\textwidth}
		\centering
		\caption{FGSM attack signature using CNN}
		{\resizebox{\textwidth}{!}{\input{Figure/CNN_sig_FG.tikz}}\label{fig:tempSe}}
	\end{subfigure}~
	\begin{subfigure}[t]{.33\textwidth}
		\centering
		\caption{FGSM attack signature using LSTM}
		{\resizebox{\textwidth}{!}{\input{Figure/LSTM_sig_FG.tikz}}\label{fig:tempS3}}
	\end{subfigure}~
	\begin{subfigure}[t]{.33\textwidth}
		\centering
		\caption{FGSM attack signature using GRU}
		{\resizebox{\textwidth}{!}{\input{Figure/GRU_sig_FG.tikz}}\label{fig:tempS4}}
	\end{subfigure}
	}
	\caption{FGSM ($\epsilon=0.3$) attack signature for sensor 2 of engine ID 49} \label{fig:FGSMSig}
\end{figure*}

\begin{figure*}[t]
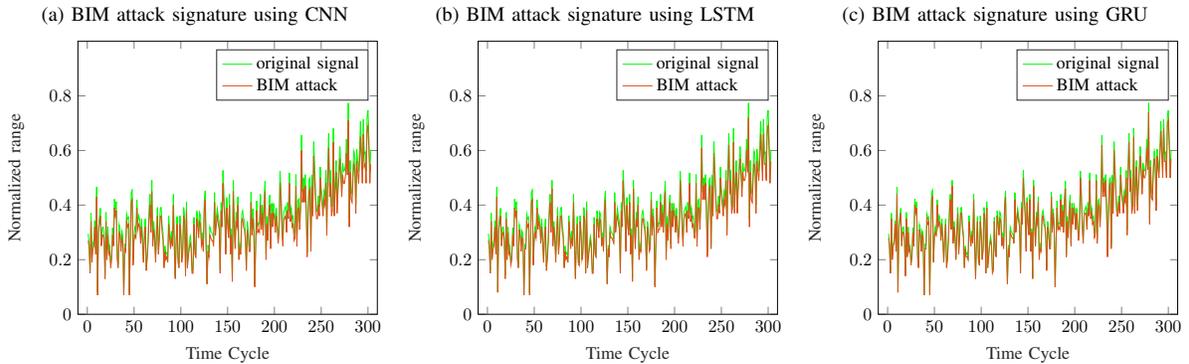

\centering
\resizebox{0.88\textwidth}{!}{
	\begin{subfigure}[t]{.33\textwidth}
		\centering
		\caption{BIM attack signature using CNN}
		{\resizebox{\textwidth}{!}{\input{Figure/CNN_sig_BI.tikz}}\label{fig:tempSe}}
	\end{subfigure}~
	\begin{subfigure}[t]{.33\textwidth}
		\centering
		\caption{BIM attack signature using LSTM}
		{\resizebox{\textwidth}{!}{\input{Figure/LSTM_sig_BI.tikz}}\label{fig:tempS3}}
	\end{subfigure}~
	\begin{subfigure}[t]{.33\textwidth}
		\centering
		\caption{BIM attack signature using GRU}
		{\resizebox{\textwidth}{!}{\input{Figure/GRU_sig_BI.tikz}}\label{fig:tempS4}}
	\end{subfigure}
	}
	\caption{BIM ($\alpha = 0.003$, $\epsilon=0.3$,~and~ $I=100$) attack signature for sensor 2 of engine ID 49} \label{fig:BIMSig}
\end{figure*}

\section{Experimental results}
In this section, we evaluate adversarial attacks by performing a case study. At first, we evaluate the performance of three PHM DL models without adversarial attacks and then apply the adversarial attacks to evaluate the impact on those models. We also evaluate the transferability property of adversarial attacks.  For the sake of reproducibility and to allow the research community to build on our findings, the artifacts (source code, datasets, etc.) of the following experiments are publicly available on our GitHub repository\footnote{https://github.com/dependable-cps/AdversarialAttack-PHM}.  
%
\subsection{Deep learning models for the turbofan engine case study}
In this work, we perform a PHM case study using an aircraft Predictive Maintenance (PdM)~\cite{zheng2018data} system. We use NASA's turbofan C-MAPSS \cite{Dataset} (Commercial Modular Aero-Propulsion System Simulation) dataset. This dataset includes 21 sensors data with a different number of operating conditions and fault conditions. More details about these 21 sensors can be found in~\cite{Dataset}. We use the FD001 sub-dataset from the dataset for our experiments. We use three DL models, specifically, LSTM, GRU, and CNN  for predicting the RUL of the aircraft engines as they are known for their applicability in the PdM domain \cite{yuan2016fault,yamak2019comparison,dong2017cnn}. Note, at a time instant, if a piece of equipment has data available for the past $N$ time cycles, then we use these data to predict the RUL at the $N+1$ time cycle. Fig. \ref{fig:LSTMStruct} shows the performance comparison of DL models with architectures LSTM(100,100,100,100) lh(80), GRU(100,100,100) lh(80), and CNN(64,64,64,64) lh(100). The notation LSTM(100,100,100,100) refers to a network that has 100 nodes in the hidden layers of the first, second, third, and fourth LSTM layers, and a sequence length of 80. In the end, there is a 1-dimensional output layer. From Fig. \ref{fig:LSTMStruct} it is evident that GRU(100, 100, 100) with a sequence length 80 provides the most accurate prediction among these three models with the least Root Mean Square Error (RMSE) of 7.62.

\subsection{Threat model for the turbofan engine PdM}
Before proceeding to the results, we describe the threat model for the turbofan engine PdM case study as follows.

\vspace{0.5em}
\noindent \textbf{Attack objective:} The objective of the attacker is to trigger either an early or delayed maintenance. An early, or in other words unnecessary maintenance can result in flight downtime and unnecessary maintenance, both of which lead to a loss of flight time, loss of human effort, loss of resources, and also incurs an extra maintenance cost. On the other hand, delayed maintenance may lead to an engine failure which might cause the loss of human lives in the worst case. 

\vspace{0.5em}
\noindent \textbf{Attack surface:} In this work, we consider both, the white-box and black-box attacks. The results of the black-box attack are demonstrated through the transferability property of the adversarial examples \cite{tramer2017space}. In the white-box attack, the adversary has access to the data and internal parameters of the DL model. In the case of aircraft predictive maintenance, the attacker can have access to the sensor data by exploiting controlled area network (CAN) bus systems aboard aircraft. The ICS-CERT published an alert on certain CAN bus systems on-board certain aircraft that might be vulnerable to cyber-threats. In this alert, an attacker with access to the aircraft could attach a malicious device to avionics CAN bus to record and inject false data. This may result in incorrect readings in an avionic equipment \cite{icsAlert}. This alert explores the possibility of capturing sensor measurements, crafting adversarial examples using the captured data, and injecting the crafted adversarial examples back into the system.

\vspace{0.5em}
\noindent \textbf{Attack scenario:} We consider an attack scenario where the adversary has access to the aircraft and could attach a device to avionics CAN bus \cite{icsAlert} as mentioned in \textit{attack surface}. The device attached to the CAN bus can record and manipulate the data. Considering a piece of equipment having $N$ time cycles and the data is transmitted through CAN to the PdM system. The DL model in the PdM system is trained to predict the RUL at the $N+1$ time cycles using the data of previous $N$ time cycles. An adversary having this knowledge can craft adversarial examples using the data of $N$ time cycles to defect the RUL prediction of $N+1$ time cycle. Once the adversarial examples are crafted, the adversary can inject these input values back to the PHM system through FDI attacks.





\vspace{0.5em}
\noindent \textbf{Attack signatures:} As mentioned earlier, the FD001 sub-dataset has 100 engines, each of which has 21 sensors. Note, 7 out of these 21 engines can be ignored since their measurements remain constant. For the rest of the 14 sensors, we used the normalization technique to convert the raw sensory data into a normalized scale. Fig. \ref{fig:3d} shows a 3-D representation of the sensor data from engine ID 49 for 300 time cycles. We use the resultant normalized dataset to generate adversarial examples using FGSM and BIM. For illustration, Fig. \ref{fig:FGSMSig} and Fig. \ref{fig:BIMSig} shows examples of a perturbed data from sensor 2 of engine ID 49 crafted using FGSM (with $\epsilon = 0.3$) and BIM (with $\alpha = 0.003$, $\epsilon = 0.3$ and $I= 100$), respectively. From Fig. \ref{fig:BIMSig} it can be observed that the BIM attack generates adversarial examples that are closer to the input. We choose $\epsilon = 0.3$ for both FGSM and BIM attacks to make sure that the crafted adversarial examples are stealthy. Such stealthy attacks often fall within the boundary conditions of the sensor measurements, and hence they are indeed hard to detect using the common attack detection mechanisms.

\begin{figure*}[t]
\centering
\resizebox{0.86\textwidth}{!}{
	\begin{subfigure}[t]{.33\textwidth}
		\centering
		\caption{CNN during FGSM (RMSE=18.35) and BIM (RMSE=31.23)}
		{\resizebox{\textwidth}{!}{
%
%
\definecolor{mycolor2}{rgb}{0.00000,0.44700,0.74100}%
\definecolor{mycolor1}{rgb}{0.85000,0.32500,0.09800}%
\definecolor{mycolor3}{rgb}{0,128,0}%
\definecolor{mycolor4}{rgb}{1.0, 0.75, 0.0}%
\begin{tikzpicture}

\begin{axis}[%
width=\textwidth,
height=2.15in,
at={(2.239in,0.602in)},
legend pos=south east,
scale only axis,
xmin=0,
xmax=38,
xlabel style={font=\color{white!15!black}},
xlabel={Engine ID},
ymin=-10,
ymax=430,
xtick ={0,5,10,15,20,25,30,35,40},
ylabel style={font=\color{white!15!black}},
ylabel={Predicted RUL in cycles},
ytick ={0,25,50,75,100,125,150,175,200,225,250,275},
axis background/.style={fill=white},
legend style={legend cell align=left, align=left,draw=white!15!black},
legend pos=north east
]
\addplot [color=mycolor2,dashed,line width = 1.5pt]
  table[row sep=crcr]{%
1	90	\\
2	94	\\
3	95	\\
4	123	\\
5	94	\\
6	49	\\
7	15	\\
8	19	\\
9	96	\\
10	89	\\
11	7	\\
12	6	\\
13	10	\\
14	9	\\
15	58	\\
16	113	\\
17	20	\\
18	28	\\
19	25	\\
20	102	\\
21	36	\\
22	20	\\
23	53	\\
24	71	\\
25	27	\\
26	7	\\
27	93	\\
28	9	\\
29	33	\\
30	7	\\
31	8	\\
32	57	\\
33	135	\\
34	37	\\
35	19	\\
36	84	\\
37	19	\\
};
\addlegendentry{True RUL}

\addplot [color=mycolor3,line width=1.5pt]
  table[row sep=crcr]{%
1	121.11835	\\
2	92.90427	\\
3	93.268105	\\
4	132.32799	\\
5	100.97358	\\
6	43.534203	\\
7	12.6685705	\\
8	19.066868	\\
9	96.62007	\\
10	88.308685	\\
11	5.572108	\\
12	5.688321	\\
13	4.1292844	\\
14	12.553413	\\
15	75.84984	\\
16	81.648865	\\
17	26.604113	\\
18	36.83136	\\
19	25.084362	\\
20	86.90913	\\
21	34.969227	\\
22	19.494314	\\
23	72.00111	\\
24	61.445213	\\
25	24.65117	\\
26	6.129045	\\
27	78.13131	\\
28	6.004906	\\
29	29.592495	\\
30	4.142157	\\
31	10.319334	\\
32	62.35981	\\
33	125.848816	\\
34	14.374789	\\
35	10.534246	\\
36	49.36004	\\
37	21.788422	\\
};
\addlegendentry{Predicted RUL}

\addplot [color=mycolor4,dashed,line width=0.75pt]
  table[row sep=crcr]{%
1	117.580666	\\
2	161.37567	\\
3	134.62988	\\
4	97.69388	\\
5	104.028145	\\
6	45.19819	\\
7	10.945969	\\
8	36.405354	\\
9	84.551186	\\
10	143.01698	\\
11	5.9028745	\\
12	14.755494	\\
13	9.545317	\\
14	29.503942	\\
15	90.59396	\\
16	84.38195	\\
17	37.220108	\\
18	45.41277	\\
19	44.975037	\\
20	150.65788	\\
21	46.28612	\\
22	32.314354	\\
23	52.417664	\\
24	68.23213	\\
25	51.558582	\\
26	15.333137	\\
27	91.71191	\\
28	3.775601	\\
29	47.830563	\\
30	6.0424705	\\
31	6.915611	\\
32	53.166756	\\
33	176.85385	\\
34	78.074196	\\
35	8.521686	\\
36	78.11325	\\
37	26.898987	\\
};
\addlegendentry{FGSM}
\addplot [color=mycolor1,line width=0.75pt]
  table[row sep=crcr]{%
1	216.40698	\\
2	119.80656	\\
3	157.02628	\\
4	78.63631	\\
5	144.58333	\\
6	48.642204	\\
7	29.739824	\\
8	23.312872	\\
9	98.27661	\\
10	122.55935	\\
11	9.147772	\\
12	26.808443	\\
13	17.70685	\\
14	18.641598	\\
15	81.31573	\\
16	170.5421	\\
17	73.75901	\\
18	26.651686	\\
19	99.78363	\\
20	90.91672	\\
21	65.6794	\\
22	22.757923	\\
23	57.511227	\\
24	48.56497	\\
25	50.688606	\\
26	4.23665	\\
27	109.87508	\\
28	28.352571	\\
29	140.20853	\\
30	6.3955336	\\
31	67.985985	\\
32	39.98313	\\
33	250.33447	\\ 
34	34.57362	\\
35	23.916103	\\
36	27.609478	\\
37	33.05631	\\
};
\addlegendentry{BIM attacked}
\end{axis}
\end{tikzpicture}%
}\label{fig:tempSe}}
	\end{subfigure}~
	\begin{subfigure}[t]{.33\textwidth}
		\centering
		\caption{LSTM during FGSM (RMSE=13.65) and BIM (RMSE=26.35)}
		{\resizebox{\textwidth}{!}{
%
%
\definecolor{mycolor2}{rgb}{0.00000,0.44700,0.74100}%
\definecolor{mycolor1}{rgb}{0.85000,0.32500,0.09800}%
\definecolor{mycolor3}{rgb}{0,128,0}%
\definecolor{mycolor4}{rgb}{1.0, 0.75, 0.0}%
\begin{tikzpicture}

\begin{axis}[%
width=\textwidth,
height=2.15in,
at={(2.239in,0.602in)},
legend pos=south east,
scale only axis,
xmin=0,
xmax=38,
xlabel style={font=\color{white!15!black}},
xlabel={Engine ID},
ymin=-5,
ymax=310,
xtick ={0,5,10,15,20,25,30,35,40},
ylabel style={font=\color{white!15!black}},
ylabel={Predicted RUL in cycles},
ytick ={0,25,50,75,100,125,150,175,200,225,250},
axis background/.style={fill=white},
legend style={legend cell align=left, align=left,draw=white!15!black},
legend pos=north east
]
\addplot [color=mycolor2,dashed,line width = 1.5pt]
  table[row sep=crcr]{%
1	90	\\
2	94	\\
3	95	\\
4	123	\\
5	94	\\
6	49	\\
7	15	\\
8	19	\\
9	96	\\
10	89	\\
11	7	\\
12	6	\\
13	10	\\
14	9	\\
15	58	\\
16	113	\\
17	20	\\
18	28	\\
19	25	\\
20	102	\\
21	36	\\
22	20	\\
23	53	\\
24	71	\\
25	27	\\
26	7	\\
27	93	\\
28	9	\\
29	33	\\
30	7	\\
31	8	\\
32	57	\\
33	135	\\
34	37	\\
35	19	\\
36	84	\\
37	19	\\
};
\addlegendentry{True RUL}

\addplot [color=mycolor3,line width=1.5pt]
  table[row sep=crcr]{%
1	105.065544	\\
2	100.61328	\\
3	78.461136	\\
4	110.335526	\\
5	97.57628	\\
6	45.01	\\
7	16.143446	\\
8	19.41734	\\
9	96.67172	\\
10	80.87248	\\
11	4.9050274	\\
12	4.363244	\\
13	8.012569	\\
14	4.8645906	\\
15	61.304733	\\
16	96.254395	\\
17	16.406265	\\
18	30.8285	\\
19	22.7753	\\
20	94.62038	\\
21	32.343967	\\
22	18.336033	\\
23	53.57677	\\
24	70.483055	\\
25	22.70794	\\
26	5.1743164	\\
27	80.737015	\\
28	4.867989	\\
29	25.499502	\\
30	4.8152127	\\
31	8.783452	\\
32	70.73162	\\
33	124.99324	\\
34	30.692822	\\
35	22.531078	\\
36	55.996403	\\
37	19.103241	\\
};
\addlegendentry{Predicted RUL}

\addplot [color=mycolor4,dashed,line width=0.75pt]
  table[row sep=crcr]{%
1	113.77835	\\
2	89.65863	\\
3	84.78672	\\
4	116.80534	\\
5	89.59128	\\
6	122.76657	\\
7	3.7453299	\\
8	11.30291	\\
9	87.930786	\\
10	73.72245	\\
11	9.643351	\\
12	0.09266138	\\
13	9.538244	\\
14	6.727546	\\
15	69.410324	\\
16	86.65205	\\
17	17.642616	\\
18	19.446827	\\
19	25.27335	\\
20	112.62812	\\
21	26.965286	\\
22	21.999172	\\
23	49.247158	\\
24	20.022964	\\
25	21.016375	\\
26	15.133078	\\
27	84.36161	\\
28	2.766296	\\
29	26.493927	\\
30	1.2347933	\\
31	2.295815	\\
32	67.48109	\\
33	84.25513	\\
34	1.4612675	\\
35	26.183228	\\
36	55.50894	\\
37	15.208866	\\
};
\addlegendentry{FGSM}

\addplot [color=mycolor1,line width=0.75pt]
  table[row sep=crcr]{%
1	142.39653	\\
2	97.14818	\\
3	57.629345	\\
4	122.539276	\\
5	137.10294	\\
6	38.037746	\\
7	7.323608	\\
8	8.827688	\\
9	82.778854	\\
10	10.611979	\\
11	8.48344	\\
12	3.0688663	\\
13	17.298481	\\
14	1.3292553	\\
15	68.30866	\\
16	129.0968	\\
17	21.252337	\\
18	58.710026	\\
19	1.881922	\\
20	65.00472	\\
21	42.309635	\\
22	14.991365	\\
23	107.76872	\\
24	4.34386	\\
25	0.8082293	\\
26	1.7845931	\\
27	12.377853	\\
28	1.3371184	\\
29	33.75468	\\
30	13.119492	\\
31	15.842922	\\
32	14.507296	\\
33	171.21268	\\
34	13.208155	\\
35	23.327522	\\
36	57.9576	\\
37	4.371319	\\
};
\addlegendentry{BIM attacked}
\end{axis}
\end{tikzpicture}%
}\label{fig:tempS3}}
	\end{subfigure}~
	\begin{subfigure}[t]{.33\textwidth}
		\centering
		\caption{GRU during FGSM (RMSE=11.22) and BIM (RMSE=25.79)}
		{\resizebox{\textwidth}{!}{
%
%
\definecolor{mycolor2}{rgb}{0.00000,0.44700,0.74100}%
\definecolor{mycolor1}{rgb}{0.85000,0.32500,0.09800}%
\definecolor{mycolor3}{rgb}{0,128,0}%
\definecolor{mycolor4}{rgb}{1.0, 0.75, 0.0}%
\begin{tikzpicture}

\begin{axis}[%
width=\textwidth,
height=2.15in,
at={(2.239in,0.602in)},
legend pos=south east,
scale only axis,
xmin=0,
xmax=38,
xlabel style={font=\color{white!15!black}},
xlabel={Engine ID},
ymin=-20,
ymax=310,
xtick ={0,5,10,15,20,25,30,35,40},
ylabel style={font=\color{white!15!black}},
ylabel={Predicted RUL in cycles},
ytick ={0,25,50,75,100,125,150,175,200,225,250},
axis background/.style={fill=white},
legend style={legend cell align=left, align=left,draw=white!15!black},
legend pos=north east
]
\addplot [color=mycolor2,dashed,line width = 1.5pt]
  table[row sep=crcr]{%
1	90	\\
2	94	\\
3	95	\\
4	123	\\
5	94	\\
6	49	\\
7	15	\\
8	19	\\
9	96	\\
10	89	\\
11	7	\\
12	6	\\
13	10	\\
14	9	\\
15	58	\\
16	113	\\
17	20	\\
18	28	\\
19	25	\\
20	102	\\
21	36	\\
22	20	\\
23	53	\\
24	71	\\
25	27	\\
26	7	\\
27	93	\\
28	9	\\
29	33	\\
30	7	\\
31	8	\\
32	57	\\
33	135	\\
34	37	\\
35	19	\\
36	84	\\
37	19	\\
};
\addlegendentry{True RUL}

\addplot [color=mycolor3,line width=1.5pt]
  table[row sep=crcr]{%
1	103.18149	\\
2	91.96608	\\
3	87.659134	\\
4	110.857735	\\
5	98.26723	\\
6	45.193714	\\
7	10.573164	\\
8	14.167494	\\
9	98.06535	\\
10	91.49428	\\
11	5.0302105	\\
12	3.4699638	\\
13	7.04605	\\
14	6.835829	\\
15	63.83042	\\
16	90.46167	\\
17	18.086512	\\
18	25.142395	\\
19	23.508507	\\
20	96.09594	\\
21	30.471752	\\
22	20.471214	\\
23	48.31953	\\
24	63.22254	\\
25	22.717337	\\
26	5.1027827	\\
27	81.48596	\\
28	4.135732	\\
29	24.22692	\\
30	4.342657	\\
31	7.7642283	\\
32	57.76628	\\
33	143.81055	\\
34	26.870972	\\
35	19.014578	\\
36	53.08556	\\
37	15.534975	\\
};
\addlegendentry{Predicted RUL}

\addplot [color=mycolor4,dashed, line width=0.75pt]
  table[row sep=crcr]{%
1	105.73299	\\
2	92.507324	\\
3	87.48141	\\
4	120.41193	\\
5	73.9154	\\
6	54.171085	\\
7	3.3296347	\\
8	16.401505	\\
9	93.89816	\\
10	63.927647	\\
11	10.642192	\\
12	3.3168173	\\
13	10.296778	\\
14	5.569535	\\
15	76.69258	\\
16	102.19404	\\
17	13.265596	\\
18	18.515049	\\
19	24.794868	\\
20	112.773315	\\
21	24.070803	\\
22	29.67625	\\
23	54.959743	\\
24	34.32243	\\
25	18.574474	\\
26	21.771688	\\
27	83.32156	\\
28	3.1031785	\\
29	27.364794	\\
30	3.4475985	\\
31	3.347395	\\
32	82.70305	\\
33	97.269165	\\
34	3.2972817	\\
35	28.469852	\\
36	52.871487	\\
37	23.027472	\\
};
\addlegendentry{FGSM}

\addplot [color=mycolor1,line width=0.75pt]
  table[row sep=crcr]{%
1	150.26213	\\
2	111.60277	\\
3	68.07716	\\
4	194.15761	\\
5	193.88094	\\
6	56.90304	\\
7	11.176004	\\
8	2.7962008	\\
9	106.95777	\\
10	4.6226306	\\
11	12.133966	\\
12	4.635399	\\
13	2.7439685	\\
14	3.2112956	\\
15	68.31919	\\
16	126.21931	\\
17	23.807533	\\
18	43.00671	\\
19	5.7348766	\\
20	65.9904	\\
21	43.641674	\\
22	2.753446	\\
23	60.175026	\\
24	2.9076605	\\
25	3.2471251	\\
26	3.3483648	\\
27	2.7866514	\\
28	3.1789038	\\
29	45.4625	\\
30	18.75372	\\
31	36.91174	\\
32	2.762103	\\
33	160.36317	\\
34	2.7808046	\\
35	28.89928	\\
36	54.947517	\\
37	10.291281	\\
};
\addlegendentry{BIM attacked}
\end{axis}
\end{tikzpicture}%
}\label{fig:tempS4}}
	\end{subfigure}
	}
	\caption{RUL estimation under FGSM ($\epsilon=0.3$) and BIM ($\alpha = 0.003$,~$\epsilon=0.3$,~and~ $I=100$) attack}\label{fig:FGSM}
\end{figure*}
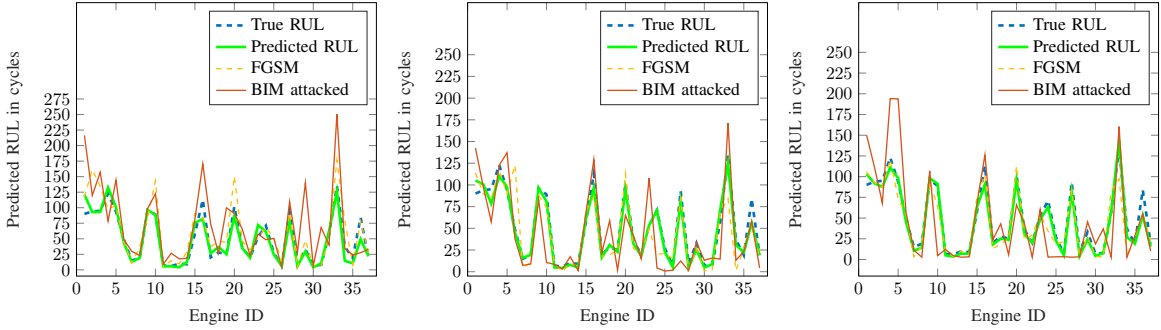
\begin{figure*}[ht]
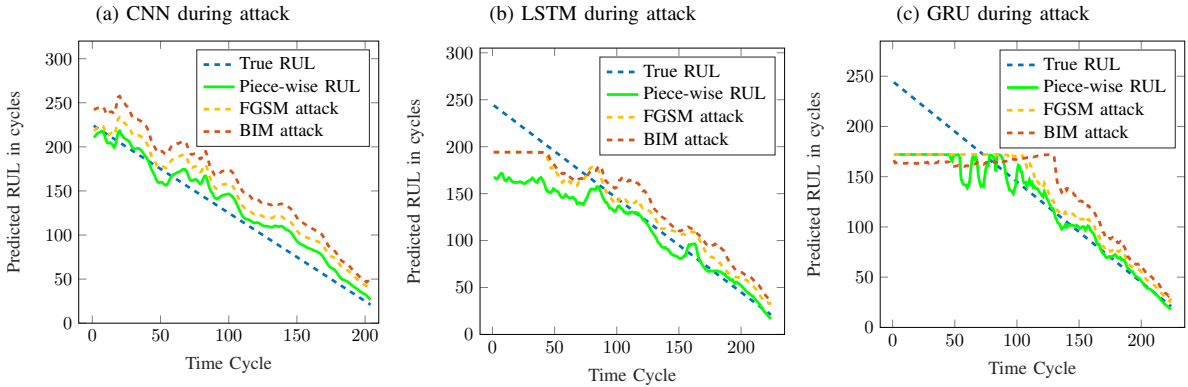

\captionsetup[subfigure]{justification=centering}
	\centering
	\resizebox{0.88\textwidth}{!}{
		\begin{subfigure}[t]{.32\textwidth}
		\caption{CNN during attack}
		{\resizebox{\textwidth}{!}{\input{Figure/CNN_FGSM_BIM_Piece.tikz}}\label{fig:tempS21}}
	\end{subfigure}~
	\begin{subfigure}[t]{.32\textwidth}
		\caption{LSTM during attack}
		{\resizebox{\textwidth}{!}{\input{Figure/LSTM_FGSM_BIM_Piece.tikz}}\label{fig:tempS22}}
	\end{subfigure}~
		\begin{subfigure}[t]{.32\textwidth}
		\caption{GRU during attack}
		{\resizebox{\textwidth}{!}{\input{Figure/GRU_FGSM_BIM_Piece.tikz}}\label{fig:tempS23}}
	\end{subfigure}
	}
	\caption{Piece-wise RUL prediction under FGSM ($\epsilon=0.3$) and BIM ($\alpha = 0.003$,~$\epsilon=0.3$,~and~$I=100$) attack}\label{fig:piecewise3}
\end{figure*}
\subsection{Impact of adversarial attacks on turbofan engine PdM}
To analyze the impact of untargeted attacks, we create a subset of test data from FD001 in which each engine has at least 150 time cycles of data. This gives us 37 engines in the FD001 dataset. This is done since engines more time cycles data helps the DL models to make more accurate RUL predictions. This gives us 37 engines in the FD001 dataset. The resultant dataset is re-evaluated using the LSTM, CNN, and GRU-based PHM models and the obtained RMSEs are 5.83, 7.92, and 5.77, respectively.

To analyze the impact of FGSM and BIM attacks on the C-MAPSS, we craft adversarial examples using Algorithm 1 and Algorithm 2 and apply them on the DL models. From \figurename~\ref{fig:FGSM}, we observe that the FGSM attack (with $\epsilon=0.3$) increases the RMSE of CNN, LSTM and GRU models by 231\%, 234\%, and 194\%, respectively, when compared to the DL models without attack. For the BIM attack (with $\alpha = 0.003$, $\epsilon = 0.3$ and $I= 100$), we also observe the similar trend, that is the RMSE for the CNN, LSTM and GRU model is increased by 394\%, 451\%, and 446\%, respectively, when compared to the DL models without attack. In all cases, as shown in \figurename~\ref{fig:FGSM}, the BIM attack results in a larger RMSE when compared to the FGSM attack.

The FGSM and BIM attacks can cause an under-prediction or over-prediction as mentioned in the attacker's objective. For instance, as shown in \figurename~\ref{fig:FGSM}, the CNN model predicts the RUL (without attack) of 125 (in hours) for engine ID 33 and 132 (in hours) for engine ID 4. After performing the FGSM and BIM attacks for engine ID 9, the same CNN model predicts the RUL (in hours) as 176 and 250, respectively. This represents a 14\% and 20\% increase in RUL after FGSM and BIM attacks. For engine ID 4, the FGSM and BIM attacks result in RUL of 97 and 78, respectively. This represents 26.51\% and 40.9\% decrease in the predicted RUL after FGSM and BIM attacks. An over-prediction, as shown in the first case, may cause delayed maintenance, whereas an under-prediction, as shown in the latter case may cause early maintenance, both of which have catastrophic consequences.

To elucidate the impact of FGSM and BIM attacks on specific engine data, we first apply the piece-wise RUL prediction (using the same DL models) for a single-engine (in this case engine ID 17) and then apply the crafted adversarial examples. The piece-wise RUL prediction gives a better visual representation of degradation (health status) in an aircraft engine. \figurename \ref{fig:piecewise3}(a), \figurename \ref{fig:piecewise3}(b), and \figurename \ref{fig:piecewise3}(c) shows the piece-wise RUL prediction using CNN, LSTM and GRU models, respectively, at each time step. From \figurename \ref{fig:piecewise3}, it is evident that as the time approaches towards the end of life, the predicted RUL (green solid line) is closer to the true RUL (blue dashes). This is because once the RUL predictions get more accurate with the increasing amount of data. 

Next, we craft adversarial examples using both FGSM and BIM for that engine (engine ID 17), apply them for piece-wise RUL prediction, and compare their impact, as shown in \figurename \ref{fig:piecewise3}. We observe that the crafted adversarial examples have a strong impact from the beginning of the RUL prediction on the CNN model when compared to the LSTM and GRU models. The piece-wise RUL prediction after the attack on the CNN model follows the same trend of the piece-wise RUL without attack, however, the attacked RUL values remain quite far from the actual prediction. On the other hand, the impact of adversarial attacks on the GRU model is interesting since we observe that the RUL remains almost constant up to 104 time cycles and 129 time cycles for FGSM and BIM attacks, respectively, then starts decreasing. Such a phenomenon is deceiving in nature as it indicates that the engine is quite healthy and may influence a `no maintenance required' decision by the maintenance engineer.  Once again, it is evident that the BIM attack has a stronger impact on piece-wise RUL prediction when compared to the FGSM attack.

\begin{figure}[h]
\centering
\includegraphics[width=0.28\textwidth]{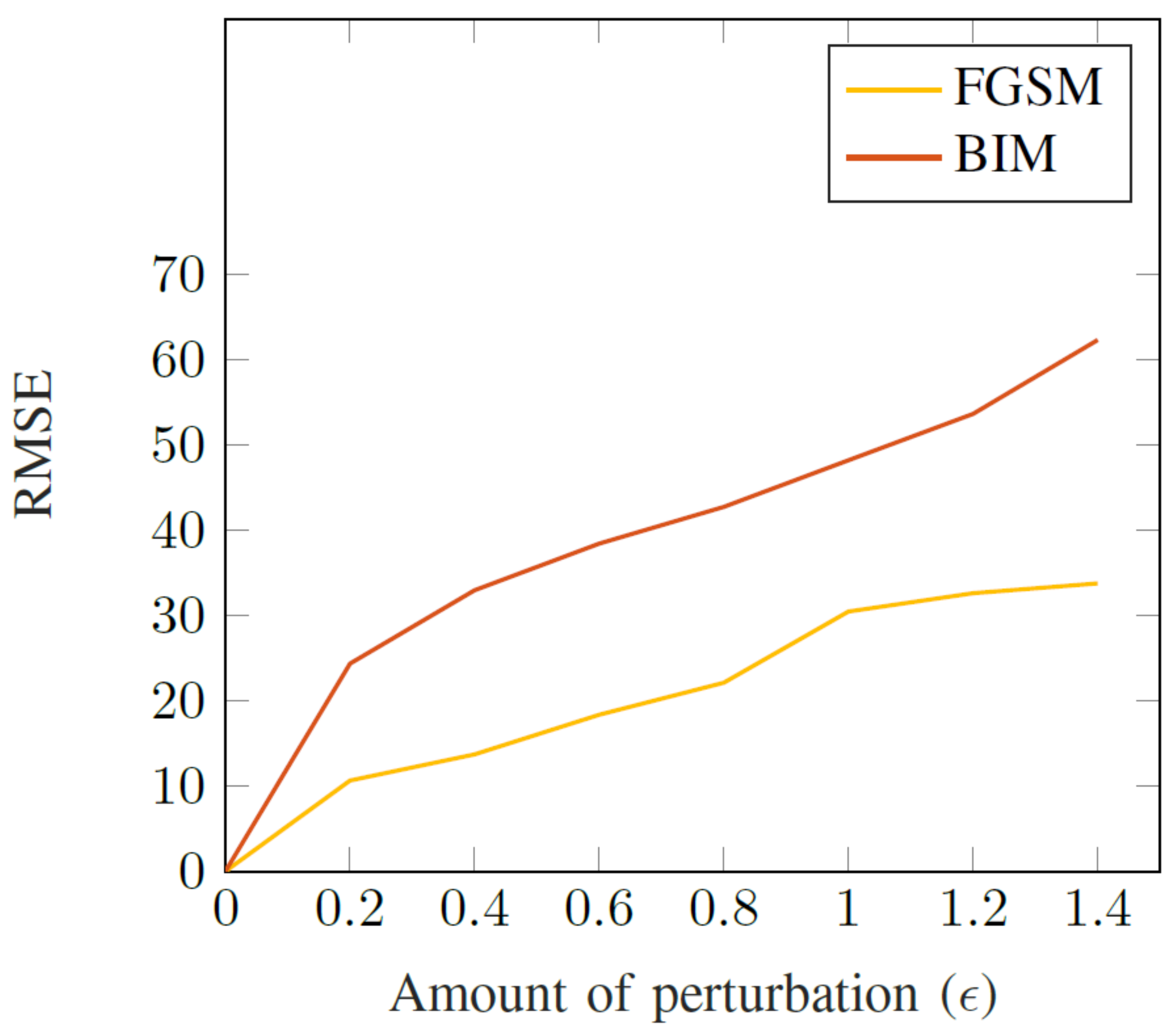}
\caption{RMSE variation with respect to the amount of perturbation ($\epsilon$) for FGSM and BIM attacks}
\label{fig:rmseEpsilon}
\end{figure}

\vspace{0.5em}
\noindent \textbf{Performance variation vs. the amount of perturbation:} In this part of the experiments, we explore the impact of the amount of perturbation $\epsilon$ on the GRU model performance in terms of RMSE. We picked the GRU model as it showed the best performance in predicting the RUL as shown in \figurename~\ref{fig:LSTMStruct}. Note, the maximum permissible value of epsilon ($\epsilon=1.4$)\cite{fawaz2019adversarial}. The obtained result is shown in \figurename~\ref{fig:rmseEpsilon}. We observe that for the larger values of $\epsilon$, the BIM attack results in higher RMSE when compared to the FGSM. For instance, for $\epsilon=1.2$, the FGSM attack results in an RMSE of 32.63, whereas the BIM attack results in an RMSE of 53.67. This shows that for the same value of $\epsilon$, BIM can generate adversarial examples impacting the RMSE approximately twice when compared to the FGSM. This is due to the fact~\cite{kurakin2016adversarial} that BIM adds a small amount of perturbation $\alpha$ on each iteration whereas FGSM adds the total amount of perturbation $\epsilon$ on each data point.

\vspace{0.3em}
\noindent \textbf{Transferability of untargeted attacks:} 
To evaluate the transferability of adversarial attacks, we apply the adversarial examples crafted for a PHM model on the other PHM models using the data of 37 engines as mentioned in the experimental setup section. As mentioned earlier, such an attack is known as the \emph{black box} attack~\cite{papernot2017practical}, where the attacker has no knowledge of the target model’s internal parameters, but still cause a considerable impact on the target model. The obtained results are shown in Table \ref{tab:BIM_Trans}. The first column (DL models) of the Table \ref{tab:BIM_Trans} represents the RMSE of the models without attack. We observe that the FGSM and BIM adversarial examples crafted for the CNN model gives a higher RMSE when transferred to other DL models. Also, another interesting fact that we observe is when transferred, adversarial examples crafted using BIM results into a higher RMSE. For instance, the CNN-based PHM model has an RMSE of 7.92. When we craft adversarial examples for the CNN model using FGSM and BIM, and transfer to the GRU model, we observe that BIM adversarial attack increases the RMSE almost two times (24.12) when compared to the FGSM (12.23). Similar trend is also observed for all other PHM models when adversarial attacks are transferred.

\begin{table}[h]
\centering
\caption {Transferability of FGSM and BIM attacks. The notation X/Y represents the RMSE after the FGSM/BIM attack}
\begin{tabular}{|c|c|c|c|}
\hline
\textbf{DL models} & \multicolumn{3}{c|}{\textbf{RMSEs after transfer}}\\ \cline{2-4}
                            & \textbf{CNN}   & \textbf{LSTM} & \textbf{GRU}                                       \\ \hline
CNN (RMSE = 7.92)   & - & 17.76 / 28.45 & 12.23 / 24.12  \\ [0.5ex] 
LSTM (RMSE = 5.83) & 18.23 / 31.13 & - & 11.33 / 18.65  \\ [0.5ex]
GRU (RMSE = 5.77)  & 16.22 / 30.45 & 10.89 / 19.52 & -  \\ [0.5ex] 
\hline
\end{tabular}
\label{tab:BIM_Trans}
\end{table}

\vspace{-2mm}	
\section{Conclusion}
In this paper, we introduce adversarial attacks to the PHM domain to reveal their vulnerabilities and show how they can be used to defect the deep learning-enabled prognostic models. We crafted adversarial examples using the FGSM and BIM algorithms for LSTM, GRU, and CNN based PHM models using NASA's turbofan engine case study. The obtained results showed that the crafted adversarial examples that include very small perturbation to the original sensor measurements can cause serious defects to the remaining useful life estimation. The BIM performed better in defecting the RUL with a smaller perturbation when compared to the FGSM. We also observed that adversarial examples crafted for CNN models are more transferable to the other DL models when compared to the LSTM and GRU. Our work is the first one to shed light on the importance of defending such adversarial attacks in the PHM domain. In the future, we would like to investigate countermeasures techniques to mitigate the adversarial threats to the PHM domain. 


\bibliographystyle{IEEEtran}

\bibliography{references}

\end{document}